\PassOptionsToPackage{table}{xcolor}
\documentclass[sigconf]{acmart}
\ifPDFTeX
  \pdfmapfile{+libertine.map}
  \pdfmapfile{+zi4.map}
  \pdfmapfile{+newtx.map}
\fi
\AtBeginDocument{%
  }

\acmConference{}{}{}
\acmBooktitle{}
\setcopyright{none}
\renewcommand{\shortauthors}{Meng et al.}
\renewcommand\footnotetextcopyrightpermission[1]{}

\usepackage{subcaption}
\usepackage{float}
\floatstyle{ruled}
\newfloat{algorithm}{tbp}{loa}
\floatname{algorithm}{Algorithm}
\usepackage{enumitem}
\usepackage{amsmath}
\usepackage{pifont}
\newcommand{\enabled}{\ding{51}}
\begin{document}

\title{Does Source-Free Time-Series Adaptation Really Need Updating the Encoder?}
\def\method{SF-FPA}

\author{\large\bfseries Shichang Meng$^{1}$, Linquan Wu$^{1}$, Xuan Ai$^{1}$, Yi Wu$^{2}$, Mingyang Liu$^{1}$, Linqi Song$^{1}$}
\affiliation{%
  \institution{$^{1}$ City University of Hong Kong\\
  $^{2}$ The Chinese University of Hong Kong}
  \city{Hong Kong}
  \country{China}}

\begin{abstract}
Source-free domain adaptation (SFDA) adapts a pretrained model to an unlabeled target domain without accessing source data, enabling the reuse of time-series predictors across sensing systems. Existing methods commonly update the encoder using unlabeled target objectives. Yet changes in users, equipment, and operating conditions also alter the spectral characteristics of input signals, raising a fundamental question: does adaptation require changing the encoder, or can we instead adapt the signal it receives? We propose Source-Free Frequency-Prototype Alignment (SF-FPA), which learns a spectral input transformation while freezing the source encoder and classifier parameters. First, a frequency adaptation layer applies amplitude and phase corrections to target signals. Separate branches adjust spectral magnitudes and phase relationships, with circular phase encoding avoiding discontinuities at the angular boundary. Second, confidence-filtered prototype alignment guides target representations toward source class prototypes. Combined with information maximization, this objective propagates gradients through the source predictor to learn spectral corrections that support class discrimination. Extensive experiments on three real-world time-series benchmarks demonstrate state-of-the-art average Macro-F1 performance. Component ablations and controlled comparisons with encoder adaptation and a parameter-matched time-domain adapter further support spectral input adaptation as an effective approach to reusing source-trained predictors. The implementation is available at \url{https://github.com/ShichangMeng/SF_FPA}.
\end{abstract}


\maketitle
\gdef\shortauthors{Meng et al.}
\fancypagestyle{standardpagestyle}{%
  \fancyhf{}
  \fancyhead[R]{\sffamily\footnotesize Meng et al.}
  \fancyfoot[C]{\footnotesize\thepage}
  \renewcommand{\headrulewidth}{0pt}
  \renewcommand{\footrulewidth}{0pt}}
\pagestyle{standardpagestyle}

\section{Introduction}
\label{sec:introduction}

Web-connected sensing services bring measurements from physical systems into applications such as activity recognition and industrial monitoring. The Web of Things (WoT) architecture supports interoperability across platforms and application domains~\cite{wotArchitecture2023}. Reusing a learned predictor across these systems introduces a further challenge: a common data interface does not ensure that measurements follow the distribution on which the predictor was trained. Changes in users, equipment, and operating conditions alter the signals presented to the model~\cite{gamboa2017deep,Ragab_2023}. For a monitoring service extended to another boiler, for example, the fault-detection task remains the same while the sensor distribution changes across physical systems~\cite{cai2021time}.

We study the learning problem underlying this reuse: how can a pretrained time-series predictor adapt to a new sensing domain when historical source records and target annotations are unavailable? Source-free domain adaptation (SFDA) addresses this constraint using source-trained models and unlabeled target measurements~\cite{SHOT2020,kim2021domain,fang2024source}. Our approach retains the encoder and classifier parameters learned from source labels and adapts the signal entering the predictor. A trainable spectral transformation connects the measurements of the target system to the source model's discriminative representations. This provides an input-level adaptation mechanism for extending sensing applications across users, operating conditions, and physical systems.

A common SFDA strategy retains the source classifier and updates the feature extractor. SHOT follows this approach through information maximization and pseudo-labeling~\cite{SHOT2020}, while other methods use self-supervision, adversarial learning, or source-free feature alignment to construct target supervision~\cite{kothandaraman2021ss,xia2021adaptive,9423431}. Subsequent methods exploit neighborhood structure and attraction--dispersion objectives to organize target representations~\cite{NRC2021,AaD2022}. Time-series approaches additionally exploit temporal and spectral structure: MAPU transfers temporal knowledge through imputation~\cite{MAPU2023}, and TFDA combines time- and frequency-domain representations~\cite{TFDA2025}. These methods provide useful supervision for adapting representations. They also motivate a complementary question: \emph{must adaptation change the learned feature extractor, or can it change the signal presented to that extractor?}

\begin{figure}[t]
    \centering
    \begin{subfigure}{0.96\columnwidth}
        \centering
        \includegraphics[width=\linewidth]{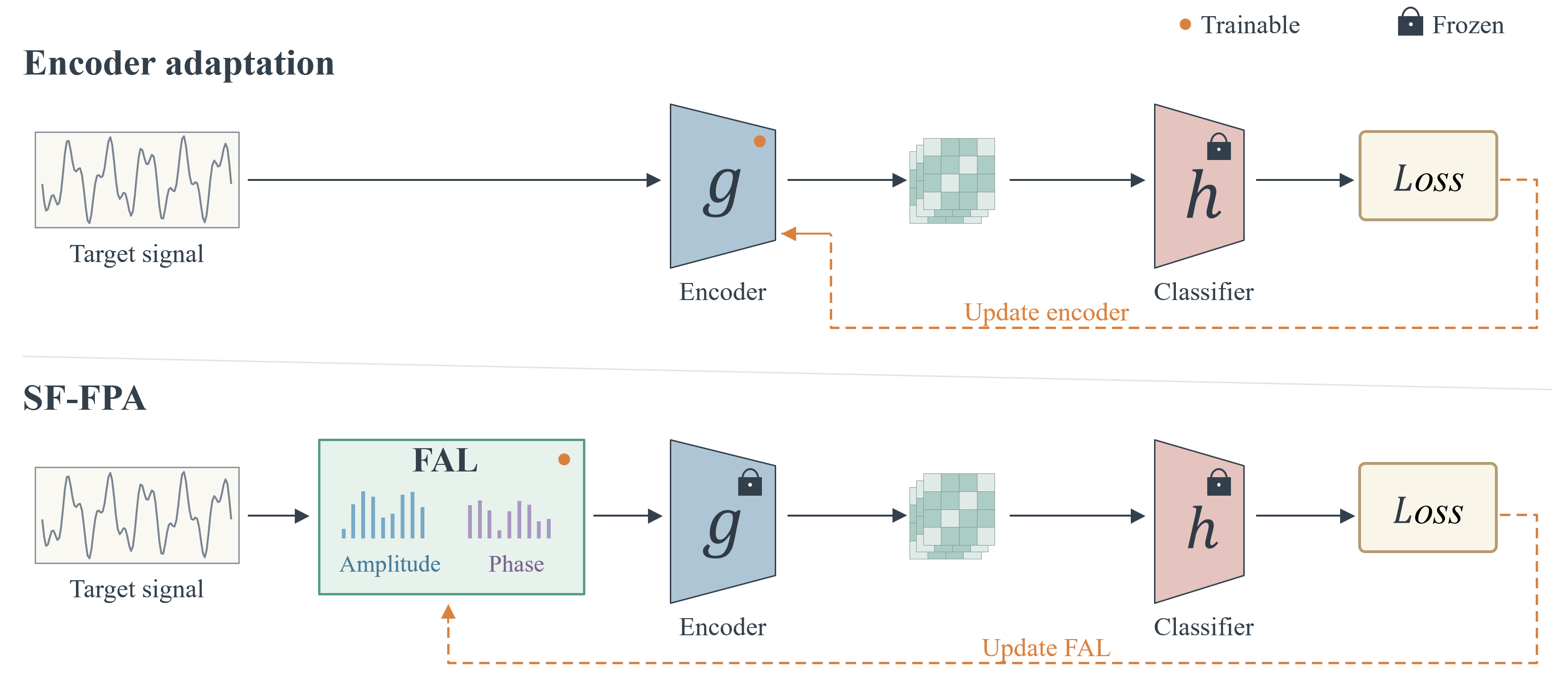}
        \caption{Encoder adaptation and spectral input adaptation.}
        \label{fig:fal_location}
    \end{subfigure}

    \vspace{0.35em}
    \begin{subfigure}{0.96\columnwidth}
        \centering
        \includegraphics[width=0.8\linewidth,height=3.5cm,keepaspectratio]{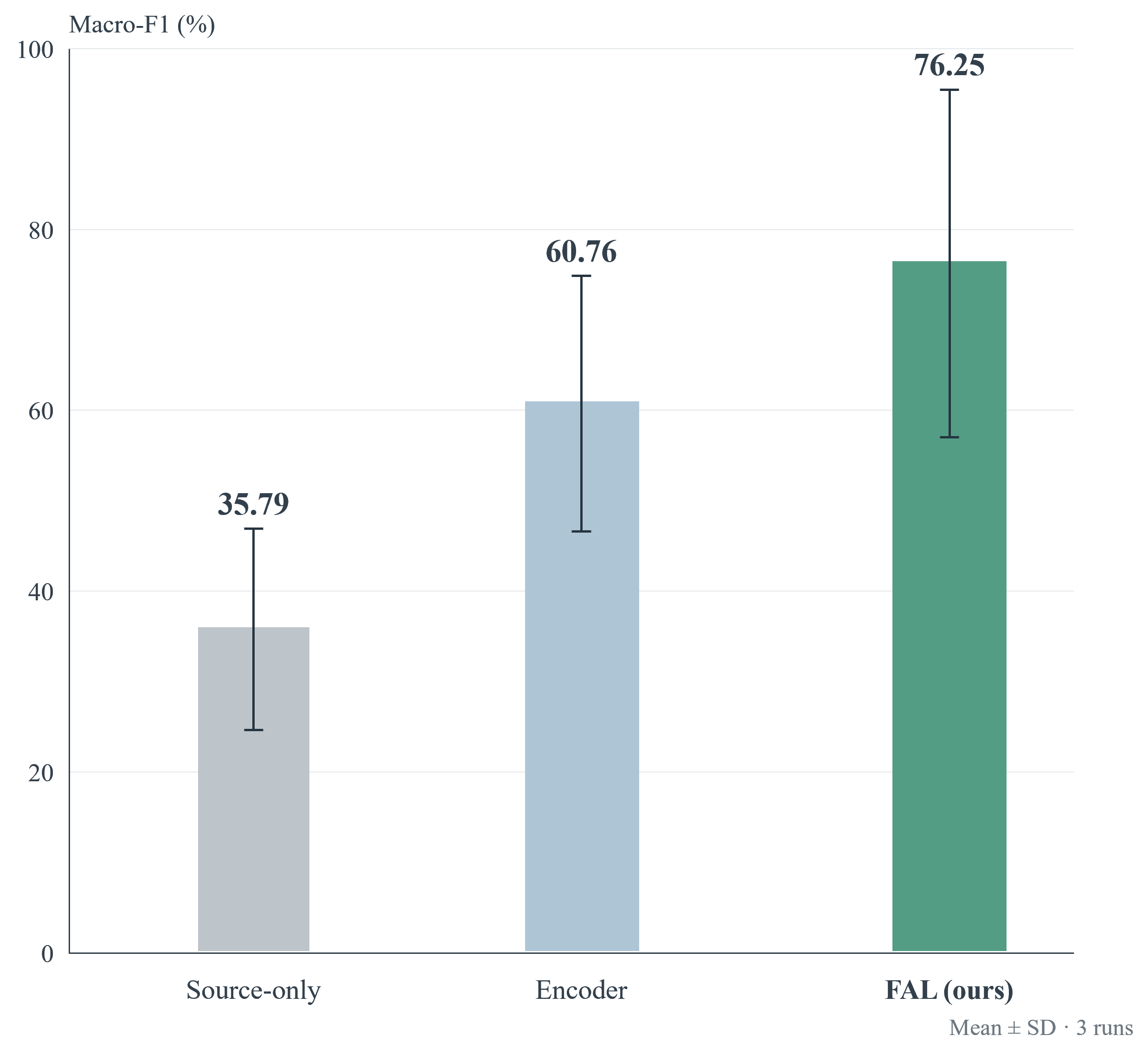}
        \caption{Adaptation-location comparison on MFD $0\!\rightarrow\!2$.}
        \label{fig:fal_result}
    \end{subfigure}
    \caption{\textbf{Adapting the signal while retaining source model parameters.}
    (a) Encoder adaptation updates the feature extractor, whereas SF-FPA learns amplitude and phase corrections through FAL with frozen source encoder and classifier parameters.
    (b) Under a shared target objective and per-task optimization settings, FAL adaptation achieves higher mean Macro-F1 on MFD $0\!\rightarrow\!2$, exceeding encoder adaptation by $15.49$ percentage points. Bars show means and error bars show sample standard deviations over three runs.}
    \label{fig:fal_motivation}
\end{figure}

The distinction matters because adapting an encoder and adapting its input place different constraints on learning. Encoder adaptation uses an unlabeled target objective to modify filters originally learned with source labels. Input adaptation instead retains those filters and learns how target measurements should excite them. Figure~\ref{fig:fal_motivation}(a) illustrates this alternative. On MFD $0\to2$, FAL reaches $76.25\%$ Macro-F1, compared with $60.76\%$ for encoder adaptation and $35.79\%$ for source-only inference (Figure~\ref{fig:fal_motivation}(b)). The $15.49$-percentage-point advantage over encoder adaptation supports a concrete design choice in this setting: concentrate learning on the input transformation while retaining the source feature-extraction parameters.

The frequency domain provides a structured way to implement this choice. Amplitude adjustments reweight frequency components, changing the spectral content available to the source filters. Phase adjustments alter how these components combine over time, changing waveform structure. Together, they provide complementary control over the signal entering the model. The adaptation variables therefore have a concrete signal-processing interpretation: they control the strength and temporal superposition of spectral components while the encoder retains its learned parameters.

Choosing those corrections requires a semantic reference. A frequency component can reflect either a domain difference or a class-discriminative pattern, so spectral manipulation alone does not determine which changes benefit recognition. We use source class feature prototypes to supply this reference. Stored during source training, these compact prototypes summarize class-specific feature directions without retaining source examples. During adaptation, confident target predictions select the corresponding prototypes, and feature alignment guides the transformed signals toward those class references. Gradients propagate through the source encoder to the input transformation: the encoder's learned filters determine how a spectral change affects the features, while prototype alignment determines which feature directions the change should favor. Thus, source knowledge guides adaptation without updating the source encoder or classifier parameters.

We instantiate this principle as \textbf{Source-Free Frequency-Prototype Alignment (SF-FPA)}. A trainable Frequency Adaptation Layer (FAL) precedes the source encoder and reconstructs each input after channel-independent amplitude and phase corrections. Bounded residual corrections and identity initialization control the transformation, while circular phase encoding handles the periodic structure of phase. Information maximization supplies prediction-level supervision, and confidence-filtered prototype alignment supplies class-level guidance. These components couple a signal-level adaptation mechanism with the semantic structure learned from source labels.

We evaluate SF-FPA on machinery fault diagnosis, human activity recognition, and boiler monitoring, covering cross-condition, cross-user, and cross-system transfer, respectively. Boiler evaluates adaptation between three physical boiler systems, grounding model reuse in an industrial sensing task. Benchmark comparisons establish recognition performance, component ablations identify the contributions of spectral branches and prototype guidance, and controlled experiments compare alternative locations and forms of adaptation across the complete WISDM benchmark. The results show that adapting the input spectrum is an effective strategy across diverse subject transfers, while the ablations confirm the complementary roles of amplitude, phase, and prototype guidance. Our contributions are summarized as follows:
\begin{itemize}
    \item We introduce a spectral input-adaptation approach to time-series SFDA that learns how target signals enter the model while retaining source encoder and classifier parameters.
    \item We develop a frequency adaptation layer with complementary amplitude and phase corrections, guided jointly by information maximization and confidence-filtered source feature prototypes.
    \item SF-FPA achieves state-of-the-art average MF1 on all three benchmarks: 87.51\% on MFD, 69.29\% on WISDM, and 70.52\% on Boiler.
\end{itemize}

\section{Related Work}
\label{sec:related}

\textbf{Cross-Domain Distribution Alignment.}
Time-series domain adaptation commonly learns transferable representations by aligning source and target distributions~\cite{wilson2020survey,liu2022deep}. CoDATS uses adversarial training for sensor representations~\cite{10.1145/3394486.3403228}, SASA aligns associative structures among variables~\cite{cai2021time}, and AdvSKM performs adversarial spectral kernel matching~\cite{ijcai2021p378}. Recurrent adversarial models also transfer temporal representations across domains~\cite{purushotham2017variational}. SLARDA incorporates autoregressive modeling and teacher--student learning~\cite{9804766}, while Raincoat addresses feature and label shifts in time-series transfer~\cite{Raincoat2023}. In rotating machinery, dynamic adversarial networks provide another source-dependent transfer strategy~\cite{WANG2025110662}. These methods exploit statistical and temporal structure with source-domain information available during adaptation. SF-FPA instead retains source knowledge in the pretrained predictor and class feature prototypes, guiding target adaptation without revisiting source data or matching two simultaneously observed domains.

\textbf{Target-Driven Self-Supervision.}
Source-free domain adaptation derives supervision from a pretrained model and unlabeled target samples, with early formulations and later surveys establishing the setting~\cite{sahoo2020unsupervised,fang2024source}. Existing methods use self-supervision, adversarial alignment, feature alignment, and information maximization~\cite{kothandaraman2021ss,xia2021adaptive,9423431,SHOT2020}. Other approaches generate source-like avatars or revisit how pretrained models should guide adaptation~\cite{qiu2021source,zhang2023rethinking}. NRC encourages prediction consistency within reciprocal neighborhoods~\cite{NRC2021}, while AaD combines neighborhood attraction with dispersion~\cite{AaD2022}. For time series, CE-SFDA uses classifier ensembles and memory-aware distillation~\cite{CESFDA2025}. Recent fault-diagnosis methods further address forgetting and uncertain pseudo-labels in source-free transfer~\cite{LI2024112672,LIU2024112443,LIU2025112475}. SF-FPA takes a different adaptation location: it optimizes a frequency adaptation layer (FAL) while freezing encoder and classifier parameters. Confidence-filtered alignment to stored source class prototypes supplies a class-specific reference alongside target prediction objectives.

\textbf{Source Knowledge Recovery.}
Another paradigm reconstructs or represents source knowledge to provide an adaptation reference. SFDA-DE estimates class-conditional source feature distributions from classifier anchors and target samples~\cite{ding2022source}. Prototypical Feature Compensation also uses source-derived class features to correct target representations~\cite{PAN2025111025}. In time-series adaptation, MAPU learns a temporal imputer during source training~\cite{MAPU2023}, while TemSR recovers source-like temporal structure through masking and recovery~\cite{TemSR2024}. These methods show the value of retaining source information in compact distributional, prototype, or temporal form. SF-FPA uses class means computed directly from labeled source features and aligns selected target features to them, giving the spectral transformation a class-specific objective without source examples or an auxiliary recovery network.

\textbf{Spectral Modeling and Input Adaptation.}
Frequency-domain learning offers structured operations for signal processing. GFNet learns global spectral filters for image recognition~\cite{GFNet2021}, and FEDformer uses frequency-enhanced blocks for forecasting~\cite{fedformer2022}. In time-series SFDA, TFDA combines temporal and frequency encoders through contrastive learning and self-distillation~\cite{TFDA2025}. Such representation-level spectral modeling does not explicitly learn how to correct a target signal before an existing source encoder processes it. Input-level spectral adaptation is explored by AIF-SFDA, which trains a frequency filter using information bottleneck and self-supervision for medical image segmentation~\cite{Li_Li_Chen_Zhong_Niu_Fu_Liu_2025}. SF-FPA develops this input-adaptation perspective for time-series classification through separate, channel-specific amplitude and phase corrections. Rather than optimizing spectral reconstruction or an information-filtering objective, it connects these corrections to source class prototypes through the encoder. This combination makes the adaptation variables signal-specific and the alignment objective class-specific, while retaining the learned encoder and classifier parameters.

\section{Methodology}
\label{sec:method}

\subsection{Problem Formulation}

Let $\mathcal{D}_s=\{(x_i^s,y_i^s)\}_{i=1}^{N_s}$
be a labeled source dataset and
$\mathcal{D}_t=\{x_j^t\}_{j=1}^{N_t}$
an unlabeled target dataset.
Each input $x\in\mathbb{R}^{C\times T}$ is a
time series with $C$ channels and $T$ time steps.
The two domains share the label space
$\mathcal{Y}=\{1,\ldots,K\}$ but follow different
joint distributions, denoted by
$\mathbb{P}_s\neq\mathbb{P}_t$.

The source model consists of a feature extractor
$g_s:\mathbb{R}^{C\times T}\rightarrow\mathbb{R}^{d}$
and a classifier
$h_s:\mathbb{R}^{d}\rightarrow\mathbb{R}^{K}$,
where $d$ is the feature dimension and $h_s$
outputs class logits.
In the source-free setting, source samples are
unavailable during target adaptation.
Only the source-trained model, stored class prototypes,
and unlabeled target samples can be accessed.

Our objective is to learn a spectral input
transformation $a_{\theta}$ that adapts target signals
to the source predictor while keeping its parameters
fixed.
The resulting class probability vector is
\begin{equation}
    \boldsymbol{\pi}_{\theta}(x)
    =
    \operatorname{softmax}
    \left(h_s\left(g_s\left(a_{\theta}(x)\right)\right)\right),
    \label{eq:target_prediction}
\end{equation}
where $\theta$ contains the adaptation parameters
and $\pi_{\theta,k}(x)$ denotes the predicted
probability of class $k$.

\subsection{Overview}

We propose \textbf{Source-Free Frequency-Prototype
Alignment} (SF-FPA), which combines spectral input
adaptation with source prototype guidance.
The framework comprises two stages.

In the source stage, we train the feature extractor
and classifier using labeled source samples and
construct a prototype for each class.
The resulting model and prototype bank retain
discriminative source information for subsequent
adaptation without storing individual source samples.

In the target stage, we introduce a Frequency
Adaptation Layer (FAL) before the source feature
extractor.
FAL learns channel-specific amplitude and phase
corrections to target signals.
Information maximization provides prediction-level
supervision, while confidence-filtered prototype
alignment supplies class-specific guidance in
feature space.
Together, these objectives optimize the spectral
transformation while preserving the source model
parameters.

\begin{figure*}[t]
    \centering
    \includegraphics[width=\textwidth]
    {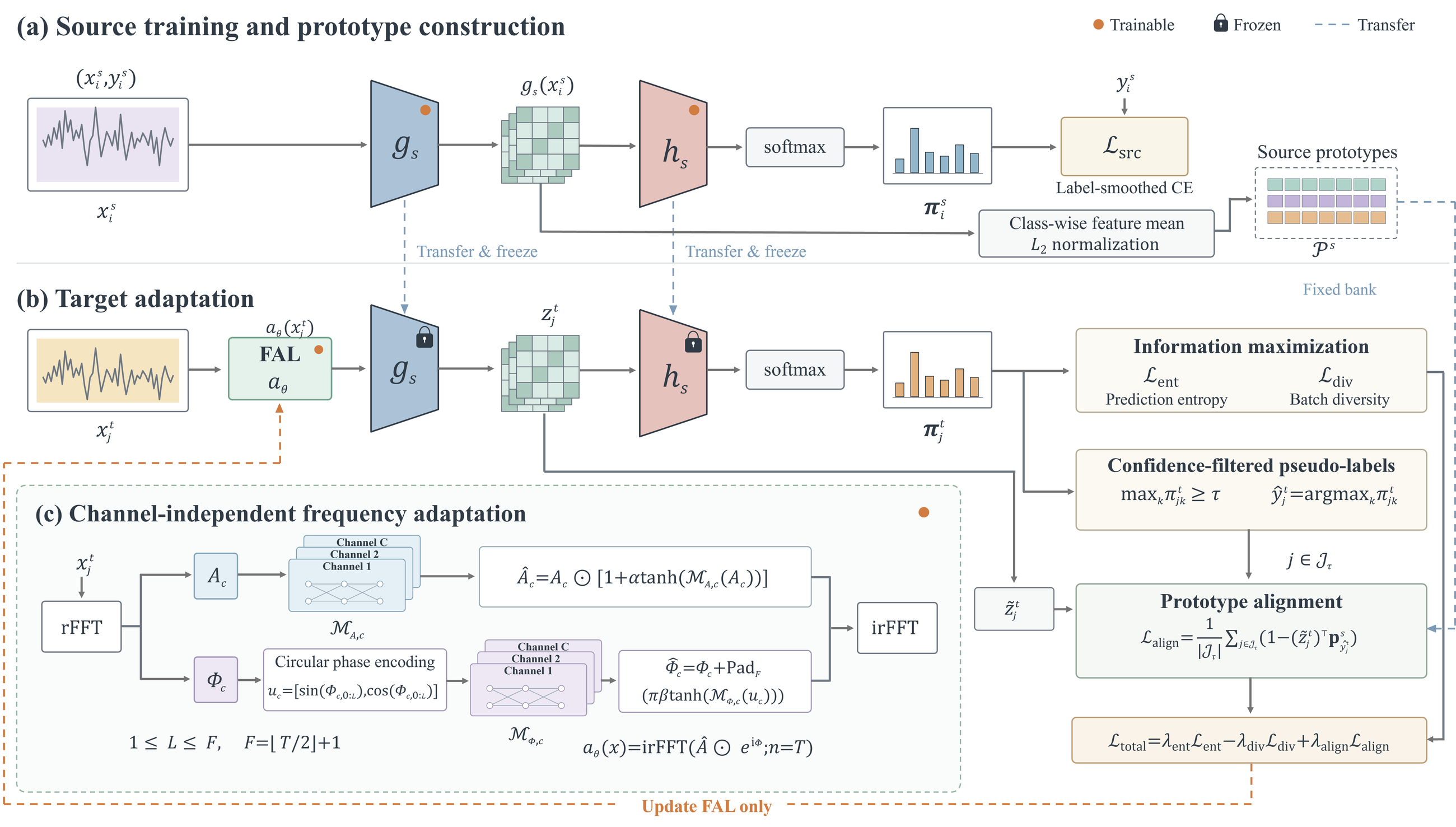}
    \caption{
    Overview of SF-FPA.
    (a) The feature extractor and classifier are
    trained on labeled source data, and class
    prototypes are constructed from source features.
    (b) The Frequency Adaptation Layer (FAL)
    adjusts the amplitude and phase of target signals
    before passing them through the source
    predictor, whose encoder and classifier parameters are frozen.
    Only FAL is optimized using information
    maximization and confidence-filtered prototype
    alignment. (c) Channel-independent amplitude and phase corrections
    followed by signal reconstruction.
    }
    \label{fig:overall}
\end{figure*}

\subsection{Source Training and Prototype Construction}
\label{sec:source_stage}

We first train $g_s$ and $h_s$ with label-smoothed
cross-entropy:
\begin{equation}
    \mathcal{L}_{\mathrm{src}}
    =
    -\frac{1}{N_s}
    \sum_{i=1}^{N_s}
    \sum_{k=1}^{K}
    \widetilde{y}_{ik}^s\log\pi_{ik}^s,
    \label{eq:source_loss}
\end{equation}
where $\pi_{ik}^s$ denotes the source predictor's probability for class $k$ on sample $i$, computed without FAL.
The smoothed label is
$\widetilde{y}_{ik}^s
=(1-\varepsilon_{\mathrm{ls}})
\mathbb{1}[y_i^s=k]+\varepsilon_{\mathrm{ls}}/K$,
with $\varepsilon_{\mathrm{ls}}=0.1$.
Here, $\mathbb{1}[\cdot]$ is the indicator function.
FAL is not applied during source training.

After training, we extract source features in
evaluation mode and construct the class prototypes:
\begin{equation}
    \begin{aligned}
        \boldsymbol{\mu}_k^s
        &=
        \frac{1}{|\mathcal{I}_k|}
        \sum_{i\in\mathcal{I}_k}g_s(x_i^s),\\
        \mathbf{p}_k^s
        &=
        \frac{\boldsymbol{\mu}_k^s}
        {\max(\|\boldsymbol{\mu}_k^s\|_2,\epsilon)},
    \end{aligned}
    \label{eq:source_prototype}
\end{equation}
where $\mathcal{I}_k=\{i:y_i^s=k\}$ is the
nonempty set of source sample indices for class $k$,
$\|\cdot\|_2$ is the Euclidean norm, and
$\epsilon>0$ ensures numerical stability.
The prototypes are stacked row-wise into
$\mathcal{P}^s\in\mathbb{R}^{K\times d}$.
This bank remains fixed during target adaptation
and provides class-specific references for target
features.

\subsection{Channel-Independent Frequency Adaptation}
\label{sec:fal}

Rather than updating the source feature extractor,
FAL adapts target samples directly in the frequency
domain.
It uses separate amplitude and phase networks for
each channel, allowing different channels to receive
different spectral corrections.

For an input $x$, we compute its one-sided spectrum
$S=\operatorname{rFFT}(x)$ and obtain its amplitude
$A=|S|$ and phase $\Phi=\operatorname{Arg}(S)$.
The number of frequency bins is
$F=\lfloor T/2\rfloor+1$.
For channel $c$, let $\mathcal{M}_{A,c}$ and
$\mathcal{M}_{\Phi,c}$ denote the amplitude and
phase networks, respectively.
Both are two-layer multilayer perceptrons with
channel-specific parameters.

\noindent\textbf{Amplitude adaptation.}
The amplitude branch predicts a bounded relative
correction and applies multiplicative modulation:
\begin{equation}
    \widehat{A}_c
    =
    A_c\odot
    \left[
        1+\alpha\tanh\left(\mathcal{M}_{A,c}(A_c)\right)
    \right],
    \label{eq:amplitude_modulation}
\end{equation}
where $\odot$ denotes elementwise multiplication
and $\alpha$ controls the correction scale.
For $0\leq\alpha\leq1$, the modulation factor
remains nonnegative.
This formulation adjusts spectral magnitudes
relative to their original values.

\noindent\textbf{Phase adaptation.}
Direct angle representations contain a discontinuity
at the $-\pi/\pi$ boundary.
We therefore encode phase using its sine and cosine,
which represent nearby angles consistently on
the unit circle.
The phase branch operates on the first $L$ frequency
bins, where $1\leq L\leq F$:
\begin{equation}
    \begin{aligned}
        u_c
        &=
        [\sin(\Phi_{c,0:L}),\cos(\Phi_{c,0:L})],\\
        \widehat{\Phi}_c
        &=
        \Phi_c+
        \operatorname{Pad}_{F}
        \left(
            \pi\beta\tanh(\mathcal{M}_{\Phi,c}(u_c))
        \right).
    \end{aligned}
    \label{eq:phase_modulation}
\end{equation}
Here, $\Phi_{c,0:L}$ contains the first $L$
phase coefficients, brackets denote concatenation,
and $\operatorname{Pad}_{F}$ appends zeros to
produce a vector of length $F$.
The scale $\beta\geq0$ bounds the selected phase
perturbations by $\pi\beta$.
When $L<F$, phase adaptation is restricted to
the lower-frequency portion of the spectrum;
$L=F$ allows all available phase bins to be adjusted.

The adapted signal is reconstructed as
\begin{equation}
    a_{\theta}(x)
    =
    \operatorname{irFFT}
    \left(
        \widehat{A}\odot
        e^{\mathrm{i}\widehat{\Phi}};\,n=T
    \right),
    \label{eq:signal_reconstruction}
\end{equation}
where $\mathrm{i}$ is the imaginary unit.
The inverse transform preserves the input sequence
length and produces a real-valued signal.
We initialize the final linear layers of both
branches to zero, so FAL initially acts as an
identity transformation up to numerical precision.

\noindent\textbf{Why adapt the input spectrum?}
Amplitude and phase provide distinct directions for changing the same signal. Multiplicative amplitude modulation changes the relative strength of its existing frequency components; phase perturbations change how those components combine over time. The two branches can therefore address different aspects of an input shift without requiring a second feature encoder. Channel independence further allows sensors with different responses to receive different corrections. The bounds control the allowed transformation range, while identity initialization starts adaptation from the source-only prediction function.

Each branch conditions its correction on the current sample's spectrum, making the transformation sample-dependent. Amplitude modulation reweights existing frequency components, and phase modulation controls their temporal superposition. Gradients through the source encoder link these signal changes to the prediction and prototype objectives. The encoder's learned filters thus supply the feature responses through which FAL is optimized, while the source class prototypes supply the desired feature directions. This gives input adaptation both a signal-level parameterization and a class-level learning target.

\subsection{Prototype-Guided Target Adaptation}
\label{sec:objectives}

For a target mini-batch of size $B$, define
$z_j^t=g_s(a_{\theta}(x_j^t))$ and
$\boldsymbol{\pi}_j^t
=\operatorname{softmax}(h_s(z_j^t))$.
Its class-$k$ probability is denoted by $\pi_{jk}^t$.
We train FAL using prediction-level information
maximization and feature-level prototype alignment.

\noindent\textbf{Information maximization.}
We minimize individual prediction entropy while
encouraging diversity in batch-averaged predictions:
\begin{equation}
    \begin{aligned}
        \mathcal{L}_{\mathrm{ent}}
        &=
        -\frac{1}{B}
        \sum_{j=1}^{B}\sum_{k=1}^{K}
        \pi_{jk}^t\log\pi_{jk}^t,\\
        \mathcal{L}_{\mathrm{div}}
        &=
        -\sum_{k=1}^{K}
        \overline{\pi}_k^t
        \log(\overline{\pi}_k^t+\epsilon_{\mathrm{div}}),
    \end{aligned}
    \label{eq:information_terms}
\end{equation}
where $\overline{\pi}_k^t
=B^{-1}\sum_{j=1}^{B}\pi_{jk}^t$ and
$\epsilon_{\mathrm{div}}>0$ is a numerical
stabilization constant.
The entropy term encourages confident predictions,
while the diversity term discourages assigning
all samples to a single class.

\noindent\textbf{Confidence-filtered prototype alignment.}
Prediction-level supervision is complemented by
the fixed source prototypes.
For each target sample, we obtain the pseudo-label
$\widehat{y}_j^t=\arg\max_k\pi_{jk}^t$ and
select it when $\max_k\pi_{jk}^t\geq\tau$,
where $\tau$ is the confidence threshold.
Let $\mathcal{J}_{\tau}$ denote the selected
sample indices and define
$\widetilde{z}_j^t
=z_j^t/\max(\|z_j^t\|_2,\epsilon)$.
The alignment objective is
\begin{equation}
    \mathcal{L}_{\mathrm{align}}
    =
    \frac{1}{|\mathcal{J}_{\tau}|}
    \sum_{j\in\mathcal{J}_{\tau}}
    \left(
        1-
        (\widetilde{z}_j^t)^\top
        \mathbf{p}_{\widehat{y}_j^t}^{s}
    \right),
    \label{eq:alignment_loss}
\end{equation}
with $\mathcal{L}_{\mathrm{align}}=0$ when
$\mathcal{J}_{\tau}$ is empty.
Pseudo-labels and sample selection are treated
as fixed during backpropagation.
This objective guides selected target features
toward their corresponding source class prototypes.

\noindent\textbf{Overall objective.}
The complete adaptation objective is
\begin{equation}
    \begin{aligned}
        \min_{\theta}\;\mathcal{L}_{\mathrm{total}}
        ={}&
        \lambda_{\mathrm{ent}}\mathcal{L}_{\mathrm{ent}}
        -
        \lambda_{\mathrm{div}}\mathcal{L}_{\mathrm{div}}\\
        &+
        \lambda_{\mathrm{align}}\mathcal{L}_{\mathrm{align}},
    \end{aligned}
    \label{eq:total_loss}
\end{equation}
where the nonnegative weights control the contributions
of the three terms.
Only FAL parameters are optimized; gradients pass
through the source predictor without updating its
parameters.
Algorithm~\ref{alg:sf_fpa} summarizes the procedure.

At inference, we apply FAL followed by the source
predictor and select the class with the highest
probability in Eq.~\eqref{eq:target_prediction}.
The prototype bank is required only during adaptation.

\begin{algorithm}[t]
\caption{Source-Free Frequency-Prototype Alignment}
\label{alg:sf_fpa}
\begin{enumerate}[leftmargin=*,nosep,label=\arabic*.]
\item Train $g_s,h_s$ on source data using $\mathcal{L}_{\mathrm{src}}$.
\item Build the fixed source class prototype bank $\mathcal{P}^s$.
\item Freeze encoder and classifier parameters; initialize FAL as identity.
\item For each target batch, adapt amplitudes and phases, reconstruct signals, and compute features and predictions.
\item Select confident samples and compute $\mathcal{L}_{\mathrm{total}}$.
\item Update only $\theta$; repeat over adaptation epochs.
\item Return $h_s\circ g_s\circ a_\theta$.
\end{enumerate}
\end{algorithm}

\section{Experiments}
\subsection{Experimental Setting}
\noindent \textbf{Benchmark Datasets.}
We evaluate three forms of sensing-domain transfer: cross-user activity recognition, cross-condition machinery diagnosis, and cross-system boiler fault detection. These benchmarks test the reuse of learned recognition parameters under changes in who generates the signal, how equipment operates, and which physical system supplies the measurements. Table~\ref{tab:dataset_stat} summarizes their characteristics.
WISDM~\cite{WISDM2011,Ragab_2023} contains smartphone accelerometer measurements from 36 subjects, with each subject defining a domain.
The Machine Fault Diagnosis (MFD) benchmark~\cite{Ragab_2023} uses Paderborn bearing vibration signals collected under four operating conditions, each treated as a separate domain.
Boiler~\cite{cai2021time} contains sensor records from three boilers collected between March 24, 2014 and November 30, 2016. Each boiler defines a domain, and the task is to detect blowdown-valve faults. This cross-boiler evaluation directly tests transfer between physical systems, a requirement for reusing fault-detection models across monitored assets.

\noindent \textbf{Compared Baselines.}
To evaluate the effectiveness of our proposed framework, we compare it with several established and recent source-free domain adaptation (SFDA) baselines with publicly available implementations:

\begin{itemize}
    \item \textbf{SHOT}~\cite{SHOT2020}: A foundational SFDA approach that freezes the source classifier and fine-tunes the target encoder by maximizing mutual information and employing self-supervised pseudo-labeling.
    
    \item \textbf{NRC}~\cite{NRC2021}: This method focuses on neighborhood reciprocity, encouraging prediction consistency among local neighbors to exploit the intrinsic structural information of the target domain.
    
    \item \textbf{AaD}~\cite{AaD2022}: It treats adaptation as an unsupervised clustering task, optimizing a contrastive-like objective that encourages neighboring features to have similar predictions while maintaining global diversity.
    
    \item \textbf{MAPU}~\cite{MAPU2023}: A temporal-imputation method that captures underlying temporal dynamics through a self-supervised temporal imputation auxiliary task.
    
    \item \textbf{TemSR}~\cite{TemSR2024}: This approach reconstructs source-like temporal dependencies via masked optimization, with local context regularization and recovery diversity maximization to mitigate domain shift.
    
    \item \textbf{CE-SFDA}~\cite{CESFDA2025}: An ensemble-based framework that enhances pseudo-label reliability through multiple classifiers and employs memory-aware distillation to preserve target data structures.
    
    \item \textbf{TFDA}~\cite{TFDA2025}: A dual-branch framework that combines temporal and frequency encoders with contrastive learning and self-distillation to enforce within- and cross-domain consistency, while using uncertainty reduction and curriculum learning to mitigate domain shift and noisy pseudo-labels.
\end{itemize}
\noindent \textbf{Implementation Details.} 
We implement SF-FPA in PyTorch using a one-dimensional convolutional neural network (CNN) as the encoder $g_s$ and a linear classifier $h_s$. Both FAL branches use channel-specific two-layer MLPs with hidden width 64. We optimize FAL with Adam while freezing encoder and classifier parameters. Appendix~\ref{appendix:hyperparameters} provides the dataset-specific learning rates, training schedules, and loss weights.

\begin{table}[htbp]
    \centering
    \caption{Dataset characteristics. Length is the number of time steps per sample. Domains correspond to subjects (WISDM), operating conditions (MFD), and individual boilers (Boiler). We evaluate nine transfers each on WISDM and MFD and three on Boiler.}
    \label{tab:dataset_stat}
    \resizebox{\linewidth}{!}{
    \begin{tabular}{l|cccc}
        \toprule
        \textbf{Dataset} & \textbf{\# Classes} & \textbf{\# Channels} & \textbf{Length} & \textbf{\# Domains} \\
        \midrule
        WISDM  & 6 & 3  & 128  & 36 \\
        MFD     & 3 & 1  & 5120 & 4  \\
        Boiler & 2 & 20 & 36   & 3  \\
        \bottomrule
    \end{tabular}}
\end{table}

\begin{table*}[t]
    \centering
    \scriptsize
    \setlength{\tabcolsep}{2pt}

    \caption{MFD MF1 (\%): mean $\pm$ standard deviation over three runs. Avg is the mean across transfers. Best and second-best means in each column are \textbf{bold} and \underline{underlined}, respectively, including ties.}
    \label{tab:fd_results}
    
    \resizebox{0.95\linewidth}{!}{ 
        \begin{tabular}{l|ccccccccc|c} 
            \toprule
            \rowcolor{red!20}
            \textbf{Algorithm} & \textbf{0$\to$1} & \textbf{0$\to$2} & \textbf{0$\to$3} & \textbf{1$\to$0} & \textbf{1$\to$3} & \textbf{2$\to$0} & \textbf{2$\to$1} & \textbf{3$\to$0} & \textbf{3$\to$2} & \textbf{Avg (\%)} \\
            \midrule
            SHOT & 71.24±21.93 & 54.96±19.25 & 71.73±22.64 & 80.37±6.55 & 98.32±0.49 & 56.92±0.16 & 81.60±7.62 & 83.92±0.60 & 68.57±0.91 & 74.18 \\
            AaD & 36.03±10.70 & 34.58±10.84 & 36.03±10.70 & 72.08±4.34 & \textbf{100.00±0.00} & 65.60±0.29 & \underline{86.57±0.00} & 74.88±4.47 & 68.50±0.00 & 63.81 \\
            NRC & 66.17±1.12 & 52.53±1.84 & 67.45±0.94 & 67.98±2.29 & 72.50±7.92 & 50.28±8.10 & 56.14±7.40 & 72.47±2.30 & 63.03±2.15 & 63.17 \\
            MAPU & 81.30±19.75 & 54.38±19.80 & 84.35±20.47 & 80.86±6.71 & 98.78±0.29 & 50.57±0.05 & 72.82±6.49 & 82.54±3.48 & 75.02±12.79 & 75.62 \\
            TemSR & 53.69±13.87 & 37.93±5.37 & 47.44±3.33 & 45.99±14.99 & 47.38±2.59 & 41.15±13.78 & 73.80±6.05 & 39.18±7.88 & 40.51±10.81 & 47.45 \\
            CE-SFDA & \underline{87.25±20.95} & \underline{72.00±24.82} & \underline{87.28±21.11} & \textbf{86.03±1.02} & \underline{99.65±0.31} & \underline{65.87±16.23} & 83.98±7.67 & \underline{86.15±2.01} & \textbf{78.98±8.79} & \underline{83.02} \\
            TFDA & 65.15±23.76 & 50.24±15.07 & 64.44±25.28 & 53.59±0.76 & 61.39±3.69 & 53.98±3.63 & 74.05±16.03 & 46.89±12.19 & 52.44±3.70 & 58.02 \\
            \midrule
            \rowcolor{blue!10}
            \textbf{Ours} & \textbf{100.00±0.00} & \textbf{76.00±20.43} & \textbf{95.83±7.22} & \underline{83.18±0.91} & \textbf{100.00±0.00} & \textbf{66.41±7.31} & \textbf{100.00±0.00} & \textbf{88.21±3.23} & \underline{77.96±9.16} & \textbf{87.51} \\
            \bottomrule
        
\end{tabular}
    }
\end{table*}

\begin{table*}[t]
    \centering
    \scriptsize
    \setlength{\tabcolsep}{2pt}

    \caption{WISDM MF1 (\%): mean $\pm$ standard deviation over three runs. Avg is the mean across transfers. Best and second-best means in each column are \textbf{bold} and \underline{underlined}, respectively, including ties.}
    \label{tab:wisdm_results}
    
    \resizebox{0.95\linewidth}{!}{ 
        \begin{tabular}{l|ccccccccc|c} 
            \toprule
            \rowcolor{red!20}
            \textbf{Algorithm} & \textbf{2$\to$32} & \textbf{7$\to$30} & \textbf{18$\to$20} & \textbf{20$\to$30} & \textbf{26$\to$2} & \textbf{9$\to$3} & \textbf{22$\to$18} & \textbf{12$\to$9} & \textbf{2$\to$20} & \textbf{Avg (\%)} \\
            \midrule
            SHOT & 76.34±2.01 & 57.10±7.76 & 55.97±9.55 & 65.80±0.09 & \underline{65.77±7.39} & 43.43±0.00 & 36.21±4.46 & \underline{76.46±0.18} & 80.54±9.43 & 61.96 \\
            AaD & 74.54±5.61 & 32.94±12.40 & 47.52±2.66 & 35.82±3.27 & 49.94±11.89 & 37.90±2.63 & 24.73±13.66 & 53.72±2.41 & 45.69±10.65 & 44.76 \\
            NRC & \textbf{78.16±5.37} & \underline{75.06±6.32} & 49.73±7.79 & 65.38±2.03 & 59.99±1.52 & \textbf{50.14±3.53} & 32.67±3.75 & 72.76±3.04 & \underline{82.00±6.71} & \underline{62.88} \\
            MAPU & 77.67±0.00 & 70.40±6.04 & 61.43±0.00 & \underline{67.83±3.26} & 58.83±1.11 & 43.40±0.06 & 32.37±3.52 & \textbf{76.57±0.18} & 73.10±0.00 & 62.40 \\
            TemSR & 62.28±9.91 & 56.23±16.29 & 58.40±1.91 & 45.84±12.71 & 52.25±7.85 & 43.18±0.43 & \underline{36.52±1.01} & 74.50±0.18 & 71.24±11.51 & 55.60 \\
            CE-SFDA & 76.78±0.77 & 60.15±2.90 & 61.45±0.30 & 65.92±1.04 & 57.00±1.99 & 41.97±2.53 & 34.94±2.05 & 75.27±2.16 & 80.38±10.62 & 61.54 \\
            TFDA & 42.66±12.91 & 53.97±25.26 & \underline{64.99±16.12} & 62.36±16.12 & 57.34±4.89 & 40.55±13.77 & 32.30±3.30 & 76.35±19.25 & 46.28±12.69 & 52.98 \\
            \midrule
            \rowcolor{blue!10}
            \textbf{Ours} & \underline{77.80±2.57} & \textbf{84.36±5.04} & \textbf{75.92±8.81} & \textbf{71.17±2.58} & \textbf{70.96±6.88} & \underline{45.56±10.40} & \textbf{37.02±4.11} & \underline{76.46±0.18} & \textbf{84.38±7.28} & \textbf{69.29} \\
            \bottomrule
        
\end{tabular}
    }
\end{table*}

\begin{table}[htbp]
    \centering
    \caption{Boiler MF1 (\%): mean $\pm$ standard deviation over three runs. Avg is the mean across transfers. Best and second-best means in each column are \textbf{bold} and \underline{underlined}, respectively, including ties.}
    \label{tab:boiler_results}
    \resizebox{\linewidth}{!}{
        \begin{tabular}{l|ccc|c} 
            \toprule
            \rowcolor{red!20}
            \textbf{Algorithm} & \textbf{1$\to$2} & \textbf{1$\to$3} & \textbf{2$\to$3} & \textbf{Avg (\%)} \\
            \midrule
            SHOT & 3.06±0.85 & 45.39±5.94 & 23.81±20.79 & 24.09 \\
            AaD & \textbf{49.52±0.00} & \underline{71.93±8.83} & 48.06±0.00 & \underline{56.50} \\
            NRC & 29.99±4.87 & 39.92±4.92 & 35.24±4.30 & 35.05 \\
            MAPU & 33.88±1.06 & 50.54±13.97 & 38.58±2.51 & 41.00 \\
            TemSR & \textbf{49.52±0.00} & 48.06±0.00 & 48.06±0.00 & 48.55 \\
            CE-SFDA & 34.76±0.19 & 44.51±2.61 & 37.34±3.92 & 38.87 \\
            TFDA & \textbf{49.52±0.01} & 52.77±8.21 & \underline{51.54±3.61} & 51.28 \\
            \midrule
            \rowcolor{blue!10}
            \textbf{Ours} & \underline{39.78±4.17} & \textbf{91.91±1.53} & \textbf{79.88±6.86} & \textbf{70.52} \\
            \bottomrule
        
\end{tabular}
    }
\end{table}

\noindent\textbf{Evaluation Protocol.}
We evaluate nine transfers on MFD, nine on WISDM, and three on Boiler, for 21 transfer tasks in total. Each table entry reports MF1 as a percentage, averaged over three runs; the accompanying standard deviation describes run-to-run variation. The dataset average is the arithmetic mean of the task means. MF1 gives equal weight to the classes, making it useful when class frequencies differ. Dataset averages summarize performance within each benchmark, rather than pooling datasets with different label spaces.

\subsection{Main Results}
\label{sec:main_results}

Tables~\ref{tab:fd_results}, \ref{tab:wisdm_results},
and \ref{tab:boiler_results} report the macro F1-score
(MF1) for each transfer task and the average across
tasks.
Results are reported as the mean and standard
deviation over three runs.
SF-FPA achieves the highest average MF1 on all three
datasets, demonstrating the effectiveness of combining
spectral input adaptation with source prototype guidance.

\noindent\textbf{Evaluation on MFD.}
As shown in Table~\ref{tab:fd_results}, SF-FPA achieves
the best performance on seven of the nine transfer
tasks, including ties, with an average MF1 of 87.51\%.
This exceeds the strongest baseline, CE-SFDA, by
4.49 percentage points.
In particular, SF-FPA achieves 100.00\% MF1 on
$0\to1$, $1\to3$, and $2\to1$.
On $2\to1$, it surpasses the best competing baseline,
AaD, by 13.43 percentage points.
These results demonstrate the effectiveness of
spectral adaptation for vibration-based fault diagnosis.
By adjusting amplitudes and phases before feature
extraction, SF-FPA provides a direct mechanism for
adapting signal characteristics while retaining the
source-trained predictor.

\noindent\textbf{Evaluation on WISDM.}
Table~\ref{tab:wisdm_results} presents the results
for cross-subject activity recognition.
SF-FPA obtains the highest MF1 on six of the nine
tasks and achieves an average of 69.29\%,
outperforming the strongest baseline, NRC, by
6.41 percentage points.
The improvement is particularly pronounced on
$18\to20$, where SF-FPA reaches 75.92\%,
exceeding the best competing result by
10.93 percentage points.
These results support the effectiveness of
channel-specific spectral corrections for
cross-subject transfer.
The amplitude branch adjusts spectral magnitudes,
while the phase branch uses a circular representation
to learn bounded phase corrections, providing
complementary ways to adapt the input signals.

\noindent\textbf{Evaluation on Boiler.}
As reported in Table~\ref{tab:boiler_results},
SF-FPA achieves an average MF1 of 70.52\%,
outperforming the strongest baseline, AaD, by
14.02 percentage points.
It obtains 91.91\% and 79.88\% on $1\to3$
and $2\to3$, exceeding the best competing results
by 19.98 and 28.34 percentage points, respectively.
The results demonstrate that the proposed framework
extends effectively to multivariate industrial signals,
using independent spectral transformations for
different sensor channels and class-specific
prototype references during adaptation.

\subsection{FAL as a Plug-in Adapter}
\label{sec:fal_baseline_extension}

To test whether spectral input adaptation complements different SFDA objectives, we add FAL to three established backbones: SHOT, AaD, and NRC. For each backbone, we use its own pretrained source checkpoint and retain its original target adaptation loss and classifier-update policy. FAL is placed before the encoder, whose parameters are frozen during adaptation. We evaluate all nine WISDM and MFD transfers and all three Boiler transfers with three seeds, selecting adapted checkpoints using the corresponding adaptation-loss criterion.

Table~\ref{tab:fal_baseline_extension} reports the mean MF1 over all transfer--seed runs for each dataset. Adding FAL improves each backbone on every benchmark. The gains range from 3.72 to 10.02 percentage points on WISDM, 6.35 to 15.80 points on MFD, and 11.23 to 34.91 points on Boiler. These consistent gains show that a trainable spectral input transformation can work alongside distinct SFDA objectives, rather than relying on a single adaptation loss.

\begin{table}[t]
\centering
\small
\setlength{\tabcolsep}{5pt}
\caption{Adding FAL to existing SFDA backbones. Entries are mean MF1 (\%) across all transfers and three seeds. Red values in parentheses show the gain over the corresponding baseline in percentage points.}
\label{tab:fal_baseline_extension}
\begin{tabular}{llrr}
\toprule
Dataset & Backbone & Baseline & +FAL \\
\midrule
WISDM & SHOT & 61.96 & 65.68 \textcolor{red}{(+3.72)} \\
      & AaD  & 44.76 & 54.78 \textcolor{red}{(+10.02)} \\
      & NRC  & 62.88 & 67.21 \textcolor{red}{(+4.34)} \\
\midrule
MFD & SHOT & 72.97 & 83.77 \textcolor{red}{(+10.80)} \\
    & AaD  & 63.81 & 79.61 \textcolor{red}{(+15.80)} \\
    & NRC  & 72.28 & 78.64 \textcolor{red}{(+6.35)} \\
\midrule
Boiler & SHOT & 40.04 & 58.16 \textcolor{red}{(+18.11)} \\
       & AaD  & 48.55 & 59.77 \textcolor{red}{(+11.23)} \\
       & NRC  & 35.49 & 70.40 \textcolor{red}{(+34.91)} \\
\bottomrule
\end{tabular}
\end{table}

\subsection{Ablation Study}
\label{sec:ablation}

We investigate the contributions of amplitude
adaptation, phase adaptation, prototype alignment,
and confidence filtering on WISDM.
Table~\ref{tab:ablation_study_wisdm} reports the
results of the source-only model, four component
ablations, and the complete SF-FPA framework.

\noindent\textbf{Amplitude and phase adaptation.}
The complete model achieves 69.29\% average MF1,
improving over the source-only model by
8.45 percentage points.
Removing amplitude adaptation reduces the score
to 66.17\%, while removing phase adaptation
reduces it to 68.35\%.
Both branches therefore contribute to the overall
performance, with amplitude adaptation providing
the larger average contribution.

\noindent\textbf{Prototype alignment.}
Removing $\mathcal{L}_{\mathrm{align}}$ reduces the
average MF1 from 69.29\% to 68.03\%.
This result supports the use of fixed source
prototypes as class-specific references for adapting
target features.

\noindent\textbf{Confidence filtering.}
Removing confidence filtering reduces the average
MF1 to 68.23\%, a decrease of 1.06 percentage points.
By restricting prototype alignment to sufficiently
confident predictions, the complete model achieves
better average performance than aligning all samples.
Together, spectral adaptation, prototype alignment,
and confidence filtering yield the highest average
MF1 among the evaluated configurations.

\begin{table}[htbp]
    \centering
    \scriptsize
    \caption{WISDM component ablations. Amp and Phase denote the spectral branches, $\mathcal{L}_{\mathrm{align}}$ denotes prototype alignment, and Conf. denotes confidence filtering. Values are average MF1 (\%) over nine transfers and three runs per transfer. Best and second-best averages are bold and underlined.}
    \label{tab:ablation_study_wisdm}
    \resizebox{0.75\linewidth}{!}{
      \begin{tabular}{cccc|c}
          \toprule
          \rowcolor{red!20}
          \textbf{Amp} & \textbf{Phase} & $\mathcal{L}_{\mathrm{align}}$ & \textbf{Conf.} & \textbf{Avg MF1 (\%)} \\
          \midrule
          \texttimes & \texttimes & \texttimes & \texttimes & 60.84 \\
          \texttimes & \enabled & \enabled & \enabled & 66.17 \\
          \enabled & \texttimes & \enabled & \enabled & \underline{68.35} \\
          \enabled & \enabled & \texttimes & \enabled & 68.03 \\
          \enabled & \enabled & \enabled & \texttimes & 68.23 \\
          \midrule
          \rowcolor{blue!10}
          \enabled & \enabled & \enabled & \enabled & \textbf{69.29} \\
          \bottomrule
      \end{tabular}
      }
\end{table}

\subsection{Controlled Evaluation of Input Adaptation}
\label{sec:controlled}

We use a controlled comparison to study where target adaptation should occur.

\noindent\textbf{Adaptation location and parameter-matched control.}
We compare five configurations across all nine WISDM transfers. \textbf{Source-only} applies the pretrained predictor directly to target samples. \textbf{Encoder} updates encoder parameters while keeping classifier parameters frozen. \textbf{FAL} learns amplitude and phase corrections before the encoder, with encoder and classifier parameters frozen. \textbf{FAL + encoder} jointly updates the adapter and encoder while keeping classifier parameters frozen. \textbf{Time-domain MLP} replaces FAL with a channel-wise residual MLP operating directly on the input sequence; its 62,835 trainable parameters closely match FAL's 62,595. All adapted configurations share the target objective. This comparison tests both the adaptation location and whether the spectral parameterization adds value beyond introducing a trainable input module. We use three random seeds and select checkpoints by adaptation loss.

\begin{figure}[t]
\centering
\includegraphics[width=0.8\columnwidth]{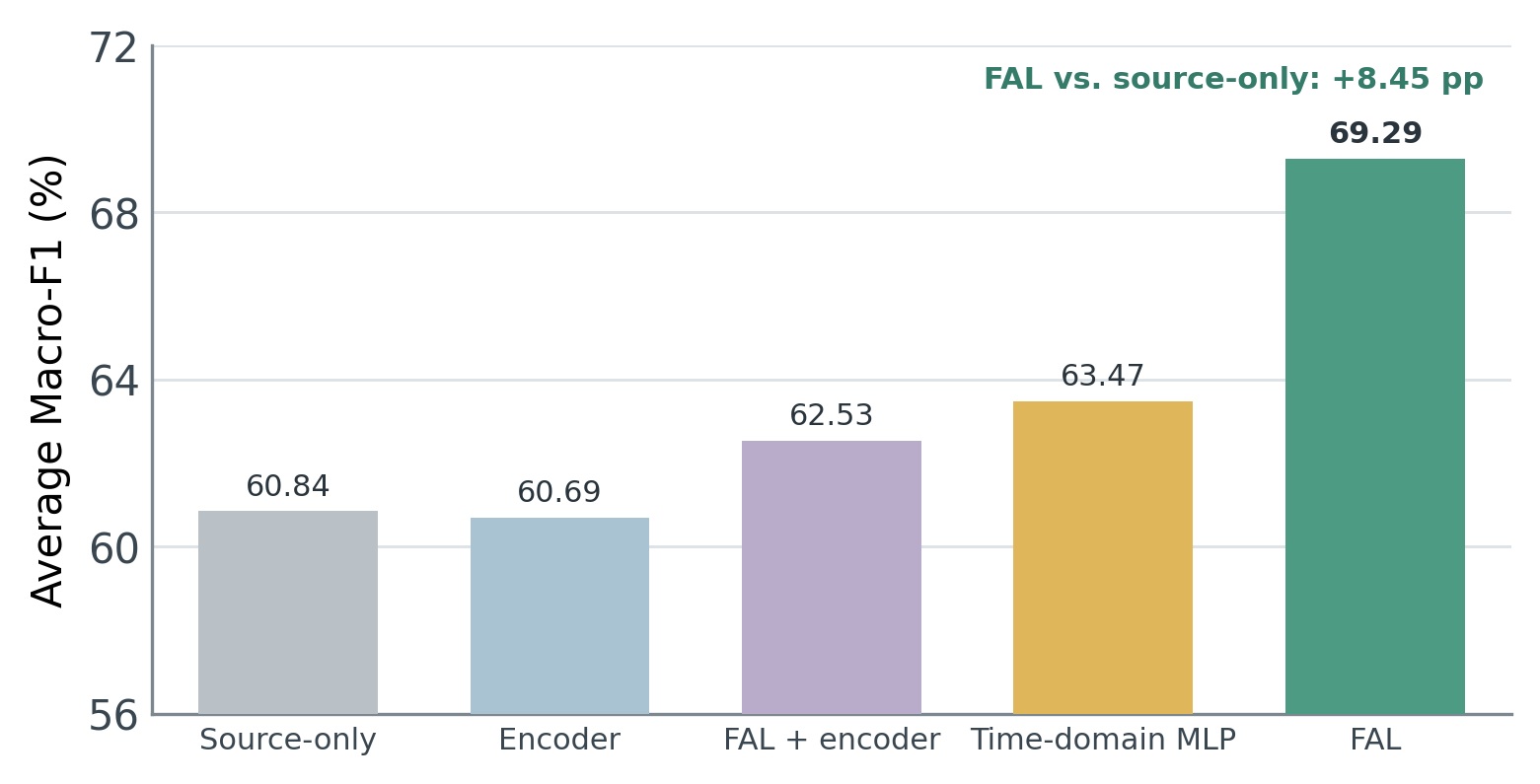}
\caption{Adaptation-location comparison across all nine WISDM transfers, averaged over three seeds. All bars use the controlled experiment protocol. labels give exact MF1 means.}
\label{fig:adaptation_location}
\end{figure}

Figure~\ref{fig:adaptation_location} reports the resulting averages: FAL reaches $69.89\%$, compared with $60.84\%$ for source-only, $60.69\%$ for encoder adaptation, $62.53\%$ for joint FAL--encoder adaptation, and $63.47\%$ for the time-domain MLP. FAL exceeds encoder adaptation by 9.20 percentage points and the parameter-matched MLP by 6.42 points. Jointly training the encoder also trails FAL by 7.36 points. Thus, expanding the set of trainable parameters does not improve this comparison; concentrating adaptation on the structured spectral transformation gives the strongest average result.

\subsection{Parameter Sensitivity Analysis}
\label{sec:sensitivity}

\begin{figure}[t]
\centering
\includegraphics[width=\columnwidth]{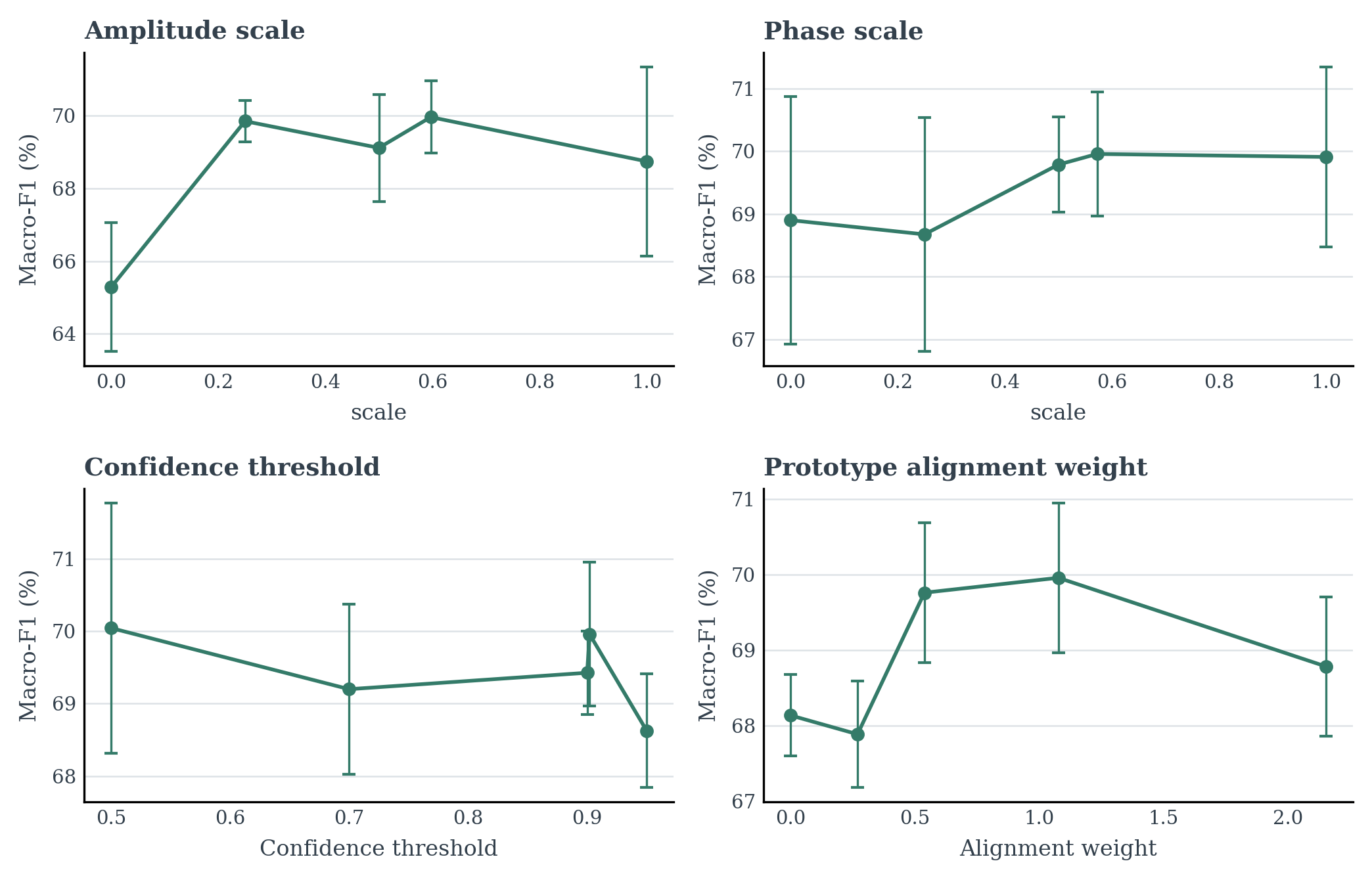}
\caption{Parameter sensitivity on all nine WISDM transfers. Panels show amplitude scale, phase scale, confidence threshold, and prototype alignment weight. Points report mean MF1 after averaging transfers within each seed; error bars show standard deviation across three seeds.}
\label{fig:sensitivity}
\end{figure}

Figure~\ref{fig:sensitivity} varies amplitude scale $\alpha$, phase scale $\beta$, confidence threshold $\tau$, and alignment weight $\lambda_{\mathrm{align}}$ one at a time across all nine WISDM transfers. For each setting, we first average MF1 over transfers within each seed, then report the mean and standard deviation over three seeds.

Nonzero amplitude scales yield 68.74--69.96\% MF1, compared with 65.28\% when amplitude adaptation is disabled. Phase scales from 0.5 to 1.0 yield 69.79--69.96\%, supporting a useful range of phase corrections. Across the tested confidence thresholds, MF1 remains between 68.63\% and 70.04\%. Alignment weights 0.538465 and 1.07693 attain 69.76\% and 69.96\%, respectively, compared with 68.13\% at zero weight. The trends favor bounded spectral corrections with a moderate prototype-alignment contribution, and show that the gains extend beyond a single parameter setting.

\subsection{Feature Visualization}
\begin{figure}[t]
\centering
\includegraphics[width=\columnwidth]{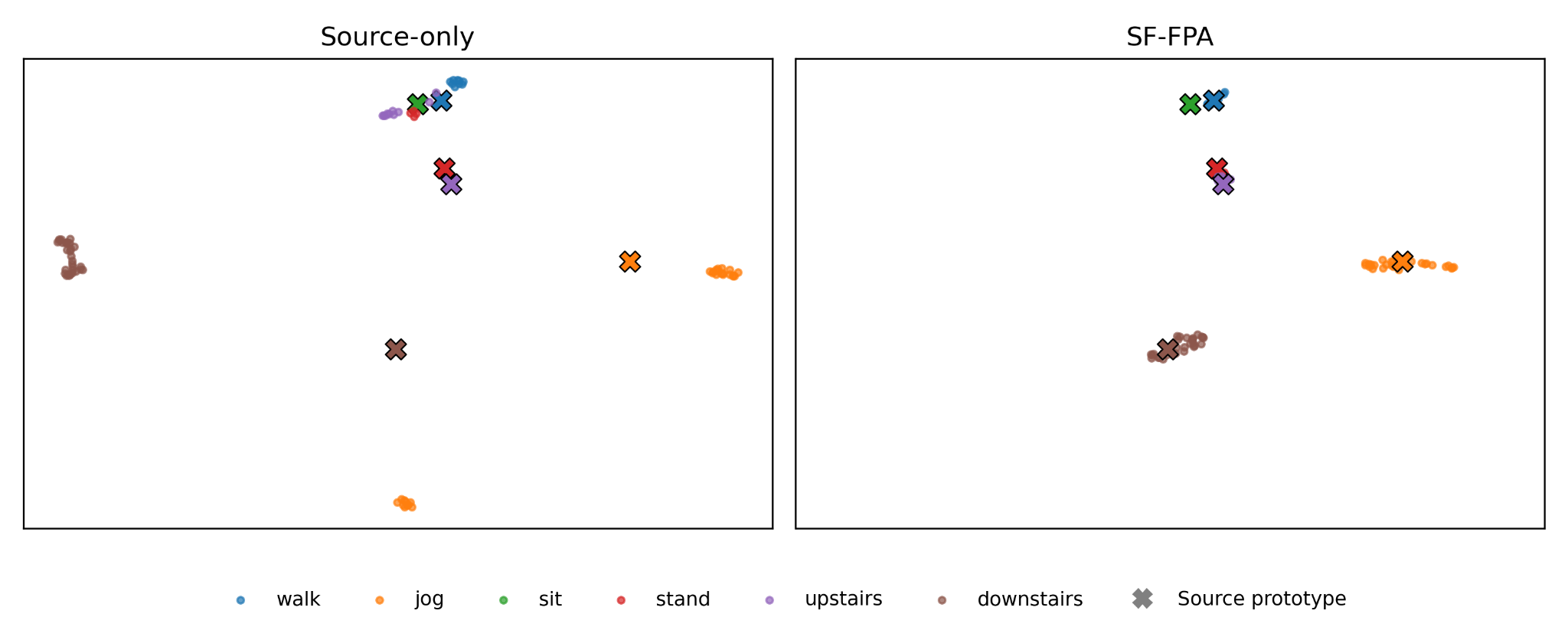}
\caption{Joint UMAP on WISDM $12\to9$ (seed 0). Left: source-only target features. Right: adapted target features. Both panels share the embedding and axis limits. Colors denote activity classes; crosses mark source class prototypes.}
\label{fig:umap_comparison}
\end{figure}

Figure~\ref{fig:umap_comparison} shows WISDM $12\to9$ using seed 0. We embed source prototypes and target features before and after adaptation together, then display the two target feature sets in the same coordinate system. After adaptation, the jogging and downstairs samples appear closer to their corresponding source prototypes in the displayed embedding. This illustrates how a transformation applied to the input can change the semantic features supplied to the classifier while retaining encoder parameters.

\section{Conclusion}
Updating encoder parameters is not a prerequisite for effective source-free time-series adaptation. SF-FPA learns how target signals enter a source-trained model through channel-independent amplitude and phase corrections. The resulting method achieves the highest average MF1 among the evaluated methods on MFD, WISDM, and Boiler. Controlled WISDM comparisons favor spectral input adaptation over encoder adaptation, joint adapter--encoder training, and a parameter-matched time-domain adapter. Ablations identify contributions from both spectral branches and prototype guidance. Together, these results establish adaptation location as a consequential design choice: source knowledge can guide a trainable input transformation while the encoder and classifier retain their learned parameters.

\clearpage
\bibliographystyle{ACM-Reference-Format}
\bibliography{sample-base}

@String{Computing = "Computing" }

@String{Computer = "{IEEE} Computer" }

@inproceedings{MAPU2023,
  title={Source-free domain adaptation with temporal imputation for time series data},
  author={Ragab, Mohamed and Eldele, Emadeldeen and Wu, Min and Foo, Chuan-Sheng and Li, Xiaoli and Chen, Zhenghua},
  booktitle={Proceedings of the 29th ACM SIGKDD conference on knowledge discovery and data mining},
  pages={1989--1998},
  year={2023}
}

@article{wilson2020survey,
  title={A survey of unsupervised deep domain adaptation},
  author={Wilson, Garrett and Cook, Diane J},
  journal={ACM Transactions on Intelligent Systems and Technology (TIST)},
  volume={11},
  number={5},
  pages={1--46},
  year={2020},
  publisher={ACM New York, NY, USA}
}

@article{WISDM2011,
  title={Activity recognition using cell phone accelerometers},
  author={Kwapisz, Jennifer R and Weiss, Gary M and Moore, Samuel A},
  journal={ACM SigKDD Explorations Newsletter},
  volume={12},
  number={2},
  pages={74--82},
  year={2011},
  publisher={ACM New York, NY, USA}
}

@inproceedings{SHOT2020,
  title={Do we really need to access the source data? source hypothesis transfer for unsupervised domain adaptation},
  author={Liang, Jian and Hu, Dapeng and Feng, Jiashi},
  booktitle={International conference on machine learning},
  pages={6028--6039},
  year={2020},
  organization={PMLR}
}

@article{NRC2021,
  title={Exploiting the intrinsic neighborhood structure for source-free domain adaptation},
  author={Yang, Shiqi and Van de Weijer, Joost and Herranz, Luis and Jui, Shangling and others},
  journal={Advances in neural information processing systems},
  volume={34},
  pages={29393--29405},
  year={2021}
}

@article{AaD2022,
  title={Attracting and dispersing: A simple approach for source-free domain adaptation},
  author={Yang, Shiqi and Jui, Shangling and Van De Weijer, Joost and others},
  journal={Advances in Neural Information Processing Systems},
  volume={35},
  pages={5802--5815},
  year={2022}
}

@article{TemSR2024,
  title={Temporal source recovery for time-series source-free unsupervised domain adaptation},
  author={Wang, Yucheng and Gong, Peiliang and Wu, Min and Ott, Felix and Li, Xiaoli and Xie, Lihua and Chen, Zhenghua},
  journal={arXiv preprint arXiv:2409.19635},
  year={2024}
}

@article{CESFDA2025,
  title={Classifier ensemble based source-free domain adaptation for time series classification},
  author={Pei, Ercheng and Zhao, Wangdong and Hu, Zhanxuan and He, Lang and Ning, Hailong and Chen, Haifeng},
  journal={Knowledge-Based Systems},
  pages={114584},
  year={2025},
  publisher={Elsevier}
}

@article{GFNet2021,
  title={Global filter networks for image classification},
  author={Rao, Yongming and Zhao, Wenliang and Zhu, Zheng and Lu, Jiwen and Zhou, Jie},
  journal={Advances in neural information processing systems},
  volume={34},
  pages={980--993},
  year={2021}
}

@inproceedings{fedformer2022,
  title={{FEDformer}: Frequency enhanced decomposed transformer for long-term series forecasting},
  author={Zhou, Tian and Ma, Ziqing and Wen, Qingsong and Wang, Xue and Sun, Liang and Jin, Rong},
  booktitle={Proc. 39th International Conference on Machine Learning (ICML 2022)},
  location = {Baltimore, Maryland},
  pages={},
  year={2022}
}

@inproceedings{Raincoat2023,
title = {Domain Adaptation for Time Series Under Feature and Label Shifts},
author = {He, Huan and Queen, Owen and Koker, Teddy and Cuevas, Consuelo and Tsiligkaridis, Theodoros and Zitnik, Marinka},
booktitle = {International Conference on Machine Learning},
year      = {2023}
}

@article{Ragab_2023,
   title={ADATIME: A Benchmarking Suite for Domain Adaptation on Time Series Data},
   volume={17},
   ISSN={1556-472X},
   url={http://dx.doi.org/10.1145/3587937},
   DOI={10.1145/3587937},
   number={8},
   journal={ACM Transactions on Knowledge Discovery from Data},
   publisher={Association for Computing Machinery (ACM)},
   author={Ragab, Mohamed and Eldele, Emadeldeen and Tan, Wee Ling and Foo, Chuan-Sheng and Chen, Zhenghua and Wu, Min and Kwoh, Chee-Keong and Li, Xiaoli},
   year={2023},
   month=may, pages={1–18} }

@inproceedings{ding2022source,
  title={Source-free domain adaptation via distribution estimation},
  author={Ding, Ning and Xu, Yixing and Tang, Yehui and Xu, Chao and Wang, Yunhe and Tao, Dacheng},
  booktitle={Proceedings of the IEEE/CVF conference on computer vision and pattern recognition},
  pages={7212--7222},
  year={2022}
}

@article{gamboa2017deep,
  title={Deep learning for time-series analysis},
  author={Gamboa, John Cristian Borges},
  journal={arXiv preprint arXiv:1701.01887},
  year={2017}
}

@article{liu2022deep,
  title={Deep unsupervised domain adaptation: A review of recent advances and perspectives},
  author={Liu, Xiaofeng and Yoo, Chaehwa and Xing, Fangxu and Oh, Hyejin and El Fakhri, Georges and Kang, Je-Won and Woo, Jonghye and others},
  journal={APSIPA Transactions on Signal and Information Processing},
  volume={11},
  number={1},
  year={2022},
  publisher={Now Publishers, Inc.}
}

@article{fang2024source,
  title={Source-free unsupervised domain adaptation: A survey},
  author={Fang, Yuqi and Yap, Pew-Thian and Lin, Weili and Zhu, Hongtu and Liu, Mingxia},
  journal={Neural Networks},
  volume={174},
  pages={106230},
  year={2024},
  publisher={Elsevier}
}

@inproceedings{cai2021time,
  title={Time series domain adaptation via sparse associative structure alignment},
  author={Cai, Ruichu and Chen, Jiawei and Li, Zijian and Chen, Wei and Zhang, Keli and Ye, Junjian and Li, Zhuozhang and Yang, Xiaoyan and Zhang, Zhenjie},
  booktitle={Proceedings of the AAAI conference on artificial intelligence},
  volume={35},
  number={8},
  pages={6859--6867},
  year={2021}
}

@article{kim2021domain,
  title={Domain adaptation without source data},
  author={Kim, Youngeun and Cho, Donghyeon and Han, Kyeongtak and Panda, Priyadarshini and Hong, Sungeun},
  journal={IEEE Transactions on Artificial Intelligence},
  volume={2},
  number={6},
  pages={508--518},
  year={2021},
  publisher={IEEE}
}

@inproceedings{kothandaraman2021ss,
  title={SS-SFDA: Self-supervised source-free domain adaptation for road segmentation in hazardous environments},
  author={Kothandaraman, Divya and Chandra, Rohan and Manocha, Dinesh},
  booktitle={Proceedings of the IEEE/CVF international conference on computer vision},
  pages={3049--3059},
  year={2021}
}

@article{qiu2021source,
  title={Source-free domain adaptation via avatar prototype generation and adaptation},
  author={Qiu, Zhen and Zhang, Yifan and Lin, Hongbin and Niu, Shuaicheng and Liu, Yanxia and Du, Qing and Tan, Mingkui},
  journal={arXiv preprint arXiv:2106.15326},
  year={2021}
}

@article{sahoo2020unsupervised,
  title={Unsupervised domain adaptation in the absence of source data},
  author={Sahoo, Roshni and Shanmugam, Divya and Guttag, John},
  journal={arXiv preprint arXiv:2007.10233},
  year={2020}
}

@INPROCEEDINGS{9423431,
  author={Yeh, Hao-Wei and Yang, Baoyao and Yuen, Pong C. and Harada, Tatsuya},
  booktitle={2021 IEEE Winter Conference on Applications of Computer Vision (WACV)}, 
  title={SoFA: Source-data-free Feature Alignment for Unsupervised Domain Adaptation}, 
  year={2021},
  volume={},
  number={},
  pages={474-483},
  doi={10.1109/WACV48630.2021.00052}}

@inproceedings{zhang2023rethinking,
  title={Rethinking the role of pre-trained networks in source-free domain adaptation},
  author={Zhang, Wenyu and Shen, Li and Foo, Chuan-Sheng},
  booktitle={Proceedings of the IEEE/CVF International Conference on Computer Vision},
  pages={18841--18851},
  year={2023}
}

@inproceedings{xia2021adaptive,
  title={Adaptive adversarial network for source-free domain adaptation},
  author={Xia, Haifeng and Zhao, Handong and Ding, Zhengming},
  booktitle={Proceedings of the IEEE/CVF international conference on computer vision},
  pages={9010--9019},
  year={2021}
}

@article{LI2024112672,
title = {Anti-forgetting source-free domain adaptation method for machine fault diagnosis},
journal = {Knowledge-Based Systems},
volume = {305},
pages = {112672},
year = {2024},
issn = {0950-7051},
doi = {https://doi.org/10.1016/j.knosys.2024.112672},
url = {https://www.sciencedirect.com/science/article/pii/S0950705124013066},
author = {Hao Li and Zongyang Liu and Jing Lin and Jinyang Jiao and Tian Zhang and Wenhao Li}
}

@article{LIU2024112443,
title = {A source free robust domain adaptation approach with pseudo-labels uncertainty estimation for rolling bearing fault diagnosis under limited sample conditions},
journal = {Knowledge-Based Systems},
volume = {304},
pages = {112443},
year = {2024},
issn = {0950-7051},
doi = {https://doi.org/10.1016/j.knosys.2024.112443},
url = {https://www.sciencedirect.com/science/article/pii/S0950705124010773},
author = {Ruiqi Liu and Wengang Ma and Feipeng Kuang and Jin Guo and Ning Zhao}
}

@inproceedings{ijcai2021p378,
  title     = {Adversarial Spectral Kernel Matching for Unsupervised Time Series Domain Adaptation},
  author    = {Liu, Qiao and Xue, Hui},
  booktitle = {Proceedings of the Thirtieth International Joint Conference on
               Artificial Intelligence, {IJCAI-21}},
  publisher = {International Joint Conferences on Artificial Intelligence Organization},
  editor    = {Zhi-Hua Zhou},
  pages     = {2744--2750},
  year      = {2021},
  month     = {8},
  note      = {Main Track},
  doi       = {10.24963/ijcai.2021/378},
  url       = {https://doi.org/10.24963/ijcai.2021/378},
}

@inproceedings{
purushotham2017variational,
title={Variational Recurrent Adversarial Deep Domain Adaptation},
author={Sanjay Purushotham and Wilka Carvalho and Tanachat Nilanon and Yan Liu},
booktitle={International Conference on Learning Representations},
year={2017},
url={https://openreview.net/forum?id=rk9eAFcxg}
}

@inproceedings{10.1145/3394486.3403228,
author = {Wilson, Garrett and Doppa, Janardhan Rao and Cook, Diane J.},
title = {Multi-Source Deep Domain Adaptation with Weak Supervision for Time-Series Sensor Data},
year = {2020},
isbn = {9781450379984},
publisher = {Association for Computing Machinery},
address = {New York, NY, USA},
url = {https://doi.org/10.1145/3394486.3403228},
doi = {10.1145/3394486.3403228},
booktitle = {Proceedings of the 26th ACM SIGKDD International Conference on Knowledge Discovery \& Data Mining},
pages = {1768–1778},
numpages = {11},
location = {Virtual Event, CA, USA},
series = {KDD '20}
}

@ARTICLE{9804766,
  author={Ragab, Mohamed and Eldele, Emadeldeen and Chen, Zhenghua and Wu, Min and Kwoh, Chee-Keong and Li, Xiaoli},
  journal={IEEE Transactions on Neural Networks and Learning Systems}, 
  title={Self-Supervised Autoregressive Domain Adaptation for Time Series Data}, 
  year={2024},
  volume={35},
  number={1},
  pages={1341-1351},
  doi={10.1109/TNNLS.2022.3183252}}

@article{PAN2025111025,
title = {Overcoming learning bias via Prototypical Feature Compensation for source-free domain adaptation},
journal = {Pattern Recognition},
volume = {158},
pages = {111025},
year = {2025},
issn = {0031-3203},
doi = {https://doi.org/10.1016/j.patcog.2024.111025},
url = {https://www.sciencedirect.com/science/article/pii/S0031320324007763},
author = {Zicheng Pan and Xiaohan Yu and Weichuan Zhang and Yongsheng Gao}
}

@article{Li_Li_Chen_Zhong_Niu_Fu_Liu_2025, title={AIF-SFDA: Autonomous Information Filter Driven Source-Free Domain Adaptation for Medical Image Segmentation}, volume={39}, url={https://ojs.aaai.org/index.php/AAAI/article/view/32498}, DOI={10.1609/aaai.v39i5.32498}, abstractNote={Decoupling domain-variant information (DVI) from domain-invariant information (DII) serves as a prominent strategy for mitigating domain shifts in the practical implementation of deep learning algorithms. However, in medical settings, concerns surrounding data collection and privacy often restrict access to both training and test data, hindering the empirical decoupling of information by existing methods. To tackle this issue, we propose an Adaptive Information Filter-driven Source-free Domain Adaptation (AIF-SFDA) algorithm, which leverages a frequency-based learnable information filter to autonomously decouple DVI and DII. Information Bottleneck (IB) and Self-supervision (SS) are incorporated to optimize the learnable frequency filter. The IB governs the information flow within the filter to diminish redundant DVI, while SS preserves DII in alignment with the specific task and image modality. Thus, the adaptive information filter can overcome domain shifts relying solely on target data. A series of experiments covering various medical image modalities and segmentation tasks were conducted to demonstrate the benefits of AIF-SFDA through comparisons with leading algorithms and ablation studies.}, number={5}, journal={Proceedings of the AAAI Conference on Artificial Intelligence}, author={Li, Haojin and Li, Heng and Chen, Jianyu and Zhong, Rihan and Niu, Ke and Fu, Huazhu and Liu, Jiang}, year={2025}, month={Apr.}, pages={4716-4724} }

@article{LIU2025112475,
title = {An adaptive source-free unsupervised domain adaptation method for mechanical fault detection},
journal = {Mechanical Systems and Signal Processing},
volume = {228},
pages = {112475},
year = {2025},
issn = {0888-3270},
doi = {https://doi.org/10.1016/j.ymssp.2025.112475},
url = {https://www.sciencedirect.com/science/article/pii/S0888327025001761},
author = {Jianing Liu and Hongrui Cao and Jaspreet Singh Dhupia and Madhurjya Dev Choudhury and Yang Fu and Siwen Chen and Jinhui Li and Bin Yv}
}

@article{WANG2025110662,
title = {A dynamic collaborative adversarial domain adaptation network for unsupervised rotating machinery fault diagnosis},
journal = {Reliability Engineering \& System Safety},
volume = {255},
pages = {110662},
year = {2025},
issn = {0951-8320},
doi = {https://doi.org/10.1016/j.ress.2024.110662},
url = {https://www.sciencedirect.com/science/article/pii/S0951832024007336},
author = {Xin Wang and Hongkai Jiang and Mingzhe Mu and Yutong Dong},
abstaract = {Acquiring sufficient fault data labels for new tasks in rotating machinery fault diagnosis is tricky. Accurately identifying faults in unlabeled scenarios is a critical and urgent practical need. Unsupervised domain adaptation (UDA) is a mainstream method to address this issue. However, most existing UDA models are static and struggle to dynamically adjust according to changes in the target task, resulting in limited diagnostic performance. To address this limitation, a dynamic collaborative adversarial domain adaptation network (DCADAN) is proposed for unsupervised rotating machinery fault diagnosis. Firstly, a multi-objective dynamic collaborative generator is designed to endow with dynamic characteristics for adjusting its own architecture, enhancing the capture capability of key domain adaptation features. Secondly, a dual-system dynamic collaborative adversarial mode is established to dynamically adjust the network training architecture, forming task-oriented refined diagnostic decision edges to steadily improve domain adaptation diagnostic capability. Finally, a multi-source domain dynamic collaborative loss is developed to match the force of multiple source domains, forming an efficient collaborative diagnostic pattern with dynamic adjustment across multi-source domains. Two case studies indicate that DCADAN demonstrates superior diagnostic performance when executing cross-domain diagnosis tasks without target labels.}
}

@article{TFDA2025,
  title={Time and frequency synergy for source-free time-series domain adaptations},
  author={Furqon, Muhammad Tanzil and Pratama, Mahardhika and Shiddiqi, Ary and Liu, Lin and Habibullah, Habibullah and Dogancay, Kutluyil},
  journal={Information Sciences},
  volume={695},
  pages={121734},
  year={2025},
  publisher={Elsevier}
}

@misc{wotArchitecture2023,
  author = {{World Wide Web Consortium}},
  title = {{Web of Things (WoT) Architecture 1.1}},
  year = {2023},
  month = dec,
  howpublished = {W3C Recommendation},
  url = {https://www.w3.org/TR/2023/REC-wot-architecture11-20231205/}
}

\clearpage
\appendix
\section{Detailed Experimental Settings}
\label{appendix:hyperparameters}

This appendix specifies the training schedules, loss coefficients, and spectral adaptation settings. 

\subsection{General Training Configurations}
\label{appendix:training_config}

Table~\ref{tab:train_params} lists the SF-FPA training settings. The main-repository baselines use the same dataset schedules. 

\begin{table}[htbp]
\centering
\caption{SF-FPA training settings by dataset.}
\label{tab:train_params}
\begin{tabular}{lccc}
\toprule
Parameter & MFD & WISDM & Boiler \\
\midrule
Adaptation epochs & 40 & 50 & 5 \\
Batch size & 32 & 64 & 32 \\
Weight decay & $10^{-4}$ & $10^{-4}$ & $10^{-4}$ \\
StepLR step size & 50 & 50 & 50 \\
StepLR decay factor & 0.5 & 0.5 & 0.5 \\
\bottomrule
\end{tabular}
\vspace{2pt}
\end{table}

\subsection{Hyperparameter Definitions}
\label{appendix:hparam_def}

\textbf{Learning rates and regularization.}
Pre-LR is the source pretraining learning rate; LR is the target adaptation learning rate. Weight decay and the StepLR schedule are given in Table~\ref{tab:train_params}.

\textbf{Prediction objective.}
Ent weights the mean target prediction entropy. IM weights the batch-mean prediction entropy in the information-maximization objective; the implementation subtracts this diversity term from the entropy-minimization loss.

\textbf{Temporal verification.}
TOV weights MAPU's temporal feature-reconstruction loss, which predicts masked temporal features from the unmasked context.

\textbf{Method-specific parameters.}
AaD's $\alpha_{\mathrm{AaD}}$ and $\beta_{\mathrm{AaD}}$ set the initial dispersion weight and its decay exponent. NRC's $\epsilon$ stabilizes the logarithm in its prediction-diversity term. Target-cls weights pseudo-label classification, and AAD weights CE-SFDA's memory-bank neighborhood loss. TemSR uses discriminator weight Disc, contrastive temperature $CL_T$, and the indicated number of temporal splits. Its remaining shared settings are source-like entropy weight 3.45814, target max--min weight 5.26417, three source-like and three target-discriminator epochs, hidden dimension 128, one masked split, anchor fraction 0.3, and disabled Gent and detach flags.

TFDA uses curriculum weights initialized as $\mu_r=1$ and $\mu_c=\mu_u=\mu_s=0.5$, with decay constants 0.005 and $10^{-4}$. The contrastive mixing coefficient is 0.5. Its adaptation objective combines pseudo-label classification, propagation, contrastive learning, uncertainty reduction, and self-distillation. Ent, IM, and TOV are therefore marked with dashes in the TFDA rows. Dashes throughout the tables indicate parameters absent from the corresponding adaptation objective.

\subsection{SF-FPA Frequency Adaptation Parameters}
\label{appendix:fal_params}

In Tables~\ref{tab:fd_hparams}--\ref{tab:boiler_hparams}, amp, phase, modes, conf, and align correspond to $\alpha$, $\beta$, $L$, $\tau$, and $\lambda_{\mathrm{align}}$, respectively. Ent and IM correspond to $\lambda_{\mathrm{ent}}$ and $\lambda_{\mathrm{div}}$. The amplitude multiplier lies in $[1-\alpha,1+\alpha]$, and each selected phase correction lies in $[-\pi\beta,\pi\beta]$. The phase branch uses the first 64 rFFT bins on MFD and WISDM and the first 18 on Boiler. Each branch has hidden width 64 and a ReLU activation; only its final linear layer's weights and biases are initialized to zero. This makes the initial spectral correction zero and the adapter an identity transformation.

For the controlled WISDM adaptation-location experiment, we use seeds 0, 1, and 2, the same source checkpoints, and the WISDM SF-FPA target hyperparameters in Table~\ref{tab:wisdm_hparams}. Adapted checkpoints are selected by the minimum per-epoch, sample-weighted mean training adaptation objective.

For each controlled comparison, source prototypes are reconstructed once from the source checkpoint and source training split, then shared by all adaptation variants. Figure~\ref{fig:adaptation_location} reports these controlled runs. Tables~\ref{tab:wisdm_results} and~\ref{tab:ablation_study_wisdm} report the benchmark and ablation runs with their saved best checkpoints.

\begin{table*}[htbp]
\centering
\caption{Algorithm-specific hyperparameters on MFD.}
\label{tab:fd_hparams}
\resizebox{\textwidth}{!}{%
\begin{tabular}{lccccc}
\toprule
Method & Pre-LR & LR & Ent & IM & Additional parameters \\
\midrule
SHOT & 0.001 & $10^{-5}$ & 0.91094 & 0.98305 & target-cls=0 \\
AaD & 0.001 & $10^{-5}$ & -- & -- & $\alpha_{\mathrm{AaD}}=1$, $\beta_{\mathrm{AaD}}=5$ \\
NRC & 0.001 & $10^{-5}$ & -- & -- & $\epsilon=10^{-5}$ \\
MAPU & 0.001 & $10^{-5}$ & 0.91094 & 0.98305 & TOV=0.169 \\
TemSR & 0.001 & $10^{-5}$ & 0.91094 & -- & Disc=50, splits=8, $CL_T=0.05$ \\
CE-SFDA & 0.001 & $10^{-5}$ & 0.91094 & 0.98305 & target-cls=1, AAD=0.01 \\
TFDA & 0.001 & $10^{-5}$ & -- & -- & Curriculum weights as defined above \\
\textbf{SF-FPA (ours)} & 0.001 & $10^{-5}$ & 0.91094 & 0.98305 & align=0.037414, amp=0.23958, phase=0.18379, modes=64, conf=0.77752 \\
\bottomrule
\end{tabular}%
}
\end{table*}

\begin{table*}[htbp]
\centering
\caption{Algorithm-specific hyperparameters on WISDM.}
\label{tab:wisdm_hparams}
\resizebox{\textwidth}{!}{%
\begin{tabular}{lccccc}
\toprule
Method & Pre-LR & LR & Ent & IM & Additional parameters \\
\midrule
SHOT & 0.001 & 0.001 & 0.41939& 0.10907 & target-cls=0.3312 \\
AaD & 0.001& 0.001 & -- & -- & $\alpha_{\mathrm{AaD}}=1$, $\beta_{\mathrm{AaD}}=10$ \\
NRC & 0.001& 0.001& -- & -- & $\epsilon=10^{-5}$ \\
MAPU & 0.001 & 0.001 & 0.41939 & 0.10907 & TOV=0.5 \\
TemSR & 0.001 & 0.001 & 0.41939 & -- & Disc=50, splits=8, $CL_T=0.05$ \\
CE-SFDA & 0.001 & 0.001 & 0.41939 & 0.10907 & target-cls=0.15, AAD=0.001 \\
TFDA & 0.001 & 0.001 & -- & -- & Curriculum weights as defined above \\
\textbf{SF-FPA (ours)} & 0.001 & 0.001 & 0.41939 & 0.10907 & align=1.07693, amp=0.59687, phase=0.57247, modes=64, conf=0.90212 \\
\bottomrule
\end{tabular}%
}
\end{table*}

\begin{table*}[htbp]
\centering
\caption{Algorithm-specific hyperparameters on Boiler.}
\label{tab:boiler_hparams}
\resizebox{\textwidth}{!}{%
\begin{tabular}{lccccc}
\toprule
Method & Pre-LR & LR & Ent & IM & Additional parameters \\
\midrule
SHOT & 0.001 & $10^{-4}$ & 0.040631 & 0.92552 & target-cls=0.3312 \\
AaD & 0.001 & $10^{-4}$ & -- & -- & $\alpha_{\mathrm{AaD}}=1$, $\beta_{\mathrm{AaD}}=10$ \\
NRC & 0.001 & $10^{-4}$ & -- & -- & $\epsilon=10^{-5}$ \\
MAPU & 0.001 & $10^{-4}$ & 0.040631 & 0.92552 & TOV=0.5 \\
TemSR & 0.001 & $10^{-4}$ & 0.040631 & -- & Disc=50, splits=4, $CL_T=0.05$ \\
CE-SFDA & 0.001 & $10^{-4}$ & 0.040631 & 0.92552 & target-cls=0.15, AAD=0.001 \\
TFDA & 0.001 & $10^{-4}$ & -- & -- & Curriculum weights as defined above \\
\textbf{SF-FPA (ours)} & 0.001 & $10^{-4}$ & 0.040631 & 0.92552 & align=0.89126, amp=0.7167, phase=0.46981, modes=18, conf=0.64167 \\
\bottomrule
\end{tabular}%
}
\vspace{2pt}
\end{table*}

\subsection{Controlled Gain and Temporal Perturbations}
\label{appendix:perturbation}

\begin{figure}[t]
\centering
\includegraphics[width=\columnwidth]{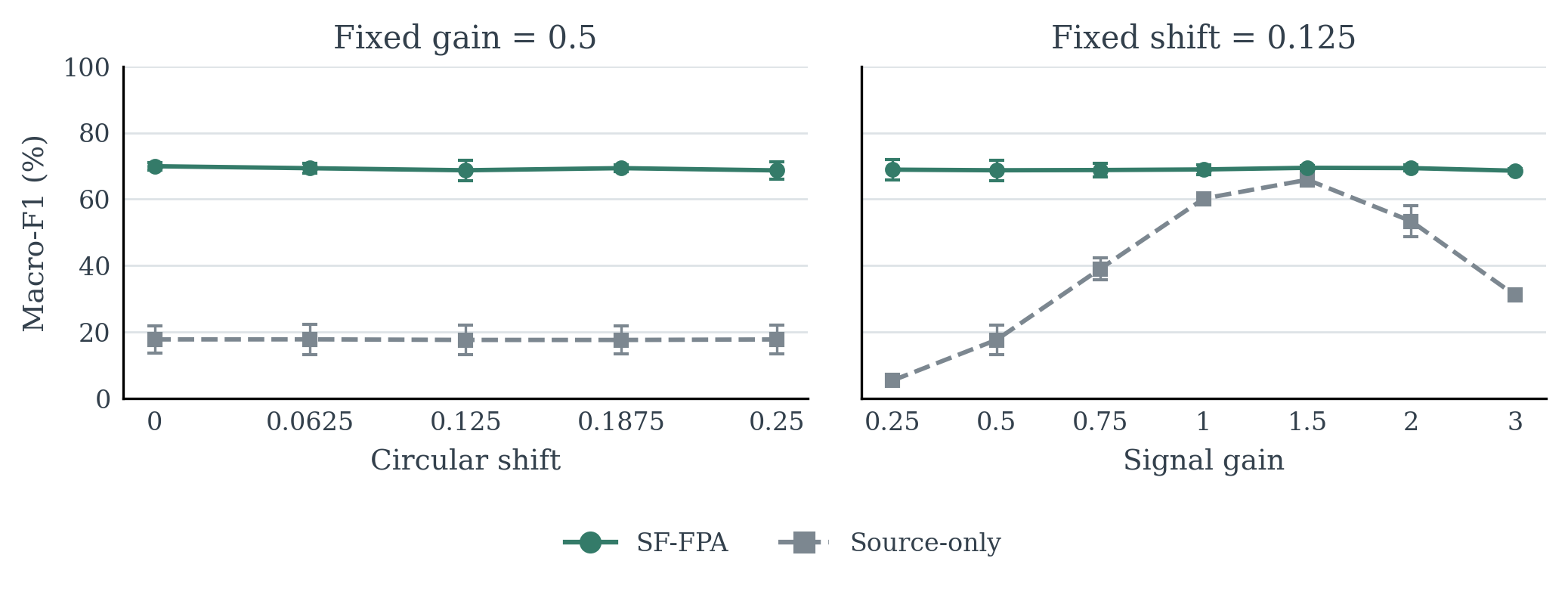}
\caption{MF1 under target gain and circular-shift perturbations, averaged across all nine WISDM transfers. Left: five shift values at fixed gain $0.5$. Right: seven gain values at fixed shift $0.125$. Each point is the mean across three seeds after averaging each seed over transfers; error bars show standard deviation.}
\label{fig:perturbation_trends}
\end{figure}

We evaluate robustness to controlled changes in target-signal magnitude and temporal alignment on all nine WISDM transfers. After the dataset's standard normalization, we multiply target training and test signals by a gain and apply a circular shift, preserving labels and sample order. Adaptation is run separately for each perturbation setting. We test gains $\{0.25, 0.5, 0.75, 1, 1.5, 2, 3\}$ and circular-shift fractions $\{0, 0.0625, 0.125, 0.1875, 0.25\}$. Figure~\ref{fig:perturbation_trends} summarizes two one-factor slices of this grid: varying shift at fixed gain $0.5$, and varying gain at fixed shift $0.125$.

At fixed gain $0.5$, SF-FPA maintains 68.68--69.93\% MF1 across the five shifts, compared with 17.65--17.84\% for source-only. At fixed shift $0.125$, SF-FPA stays within 68.59--69.45\% across all seven gains, while source-only ranges from 5.48\% to 65.86\%. The scaling is applied after normalization, so this experiment tests adaptation to deliberate magnitude changes in the normalized target inputs.

\end{document}